\documentclass[default,iicol]{sn-jnl}

\usepackage{framed,multirow}
\usepackage{algorithm}
\usepackage{algpseudocode}
\usepackage{algorithmicx}
\usepackage{amsmath,amssymb,amsfonts}
\usepackage{graphicx}
\usepackage{caption}
\usepackage{subcaption}
\usepackage{booktabs}

\jyear{2026}%

\theoremstyle{thmstyleone}%

\theoremstyle{thmstyletwo}%

\begin{document}

\title[Article Title]{Cross-Domain Generalization Limits of Vision Foundation Models in Facial Deepfake Detection}

\author[1]{\fnm{İbrahim} \sur{Delibaşoğlu}}\email{ibrahimdelibasoglu@sakarya.edu.tr}

\affil[1]{\orgdiv{Department of Software Engineering, Faculty of Computer and Information Sciences}, \orgname{Sakarya University}, \orgaddress{\street{Esentepe}, \city{Sakarya}, \postcode{54050} \country{Türkiye}}}

\abstract{The rapid evolution of generative models has enabled the creation of hyper-realistic facial deepfakes, exposing a critical vulnerability in modern digital forensics: the inability of detectors to generalize to unseen manipulation techniques. Traditional networks suffer from representation collapse, overfitting to localized artifact fingerprints of specific training generators. This work investigates whether modern Vision Foundation Models can serve as generalizable, out-of-the-box feature extractors capable of tracking forensic anomalies across entirely unseen generative manifolds. We conduct a systematic cross-domain evaluation comparing three foundational learning paradigms: fully supervised macro-semantic features (RoPE-ViT), pure self-supervised geometric features (DINOv3), and multi-teacher agglomerative representations (NVIDIA C-RADIOv4-H). By deploying frozen backbones subjected to downstream linear probing, we map the performance limitations of these architectures on the challenging DF40 benchmark. Our empirical findings expose the intrinsic trade-offs between pre-training paradigms and parameter scale, proving that while foundation models retain high discriminative capabilities for entire face synthesis, localized face editing techniques expose fundamental boundaries in linear probe evaluation structures. Source code and model weights are available in \url{http://github.com/mribrahim/deepfake}}


\maketitle

\section{Introduction}

The rapid advancement of deep generative models, particularly Denoising Diffusion Probabilistic Models (DDPMs) and Generative Adversarial Networks (GANs), has democratized the creation of hyper-realistic facial forgeries. While these technologies offer creative potential, their misuse in generating "deepfakes" poses existential threats to digital trust, social discourse, and biometric authentication integrity. Traditional detection paradigms have predominantly focused on capturing low-level pixel inconsistencies or specific architectural "fingerprints" left by generators. However, these methods suffer from a catastrophic drop in performance when confronted with domain shift. Detectors trained on GAN-based synthesis often fail to recognize artifacts from modern Diffusion-based models like CollabDiff \cite{huang2023collaborative} or DALL-E. This generalization gap stems from the fact that most detectors overfit to the specific sampling noise of the training generator rather than learning a robust representation of "realness."

In this work, we address the generalization challenge by evaluating the descriptive capacity of frozen Vision Foundation Models (VFMs) for downstream digital forensics. We hypothesize that the pre-training paradigm—whether fully supervised classification, self-supervised representation learning, or multi-teacher feature consolidation—dictates a model's intrinsic sensitivity to global structural deformations versus microscopic synthesis anomalies. By testing a curated collection of diverse frozen visual backbones, including a supervised RoPE-ViT \cite{heo2024rotary} baseline, Meta's self-supervised DINOv3 \cite{simeoni2025dinov3}, and NVIDIA's C-RADIOv4-H \cite{ranzinger2026c}, we observe how general-purpose features adapt to fine-grained forgery classification. Crucially, we explore these capabilities via a downstream linear tuning strategy. Our extensive evaluation on the DF40 benchmark~\cite{yan2024df40} maps the boundaries of cross-domain validation, clarifying how raw model capacity scales against the underlying representation chemistry of the pre-training dataset.

\section{Related Work}

Early research in deepfake detection predominantly relied on convolutional neural networks (CNNs) with standard backbones, such as Xception or EfficientNet. While these architectures achieved near-perfect accuracy on in-distribution datasets, they exhibited significant performance degradation when confronted with unseen manipulation techniques \cite{ojha2023towards}. Prior work has established baselines with such architectures; for instance, Şafak et al. \cite{csafak2024detection} evaluated the performance of lightweight CNNs, including MobileNet and EfficientNet, for detecting GAN-generated faces and subsequently employed ensemble methods. However, their study did not explicitly probe the out-of-distribution generalization limits of these pretrained models—an issue central to more recent research. Consistent with the broader literature on generalization, our experiments confirm that even models with strong pretrained initializations fail to maintain robust performance when tested on the subtle and varied artifacts produced by different, unseen deepfake methods.

To address this generalization challenge, recent studies have shifted towards proactive strategies utilizing large-scale foundational models. Ojha et al. \cite{ojha2023towards} proposed leveraging the feature space of vision-language models like CLIP, hypothesizing that frozen, general-purpose representations are more robust against manipulation-specific attacks than forgery-specialized networks. Concurrently, the advent of self-supervised learning has opened new avenues. Notably, Huang et al. \cite{huang2025rethinking} demonstrated that large-scale Vision Transformers (ViTs) pre-trained on extensive natural image datasets provide powerful representation baselines for detecting structural anomalies across various generators via simple linear probing. This foundational baseline was expanded by the introduction of DINOv3 \cite{simeoni2025dinov3}, which enhances spatial awareness and dense feature map sharpness through Gram anchoring, offering a highly resilient latent space for out-of-distribution evaluation tasks.

Spectral analysis remains a cornerstone of digital forensics, as generative models often introduce telltale periodic artifacts—for instance, from upsampling operations or convolutional checkerboard patterns. Approaches such as FreqNet \cite{tan2024frequency} explicitly employ frequency-aware learning mechanisms to capture these spectral footprints. Lanzino et al. \cite{Lanzino_2024_CVPR} enhanced robustness against compression by integrating Fast Fourier Transforms (FFT) with Local Binary Patterns (LBP) using Binary Neural Networks (BNN). While frequency models target micro-anomalies, foundation models focus on macro-structures. In this paper, we bridge these viewpoints by assessing whether multi-teacher architectures like NVIDIA's C-RADIOv4-H , which compresses semantic (DINOv3), language (SigLIP2 \cite{tschannen2025siglip2multilingualvisionlanguage}), and edge-segmentation (SAM3 \cite{carion2025sam3segmentconcepts}) features simultaneously, can naturally capture both macro-structural distortions and pixel-level boundaries within a unified, frozen feature space.

\section{Methodology}
\label{sec:methodology}

Our methodology investigates the descriptive utility of frozen feature maps generated by disparate Vision Foundation Model paradigms for downstream deepfake classification. Given an input facial Region of Interest (ROI), we extract localized spatial representations using a selected backbone, freeze its internal weights to isolate the raw representation space, and pass the resulting vectors through a lightweight linear probing framework.

\begin{figure}[htbp]
    \centering
    \begin{minipage}[b]{0.26\textwidth}
        \centering
        \includegraphics[width=\textwidth]{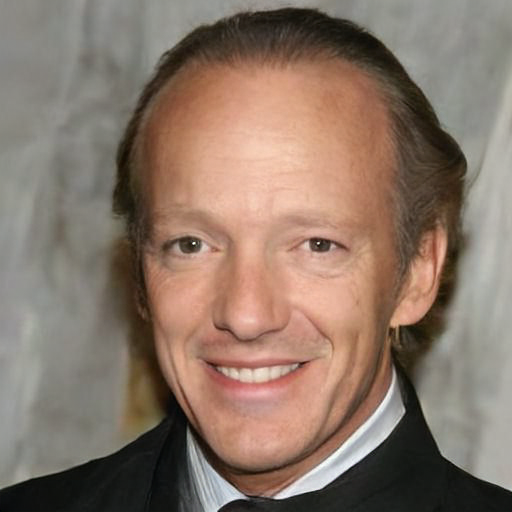}
    \end{minipage}
    \hfill
    \begin{minipage}[b]{0.18\textwidth}
        \centering
        \includegraphics[width=\textwidth]{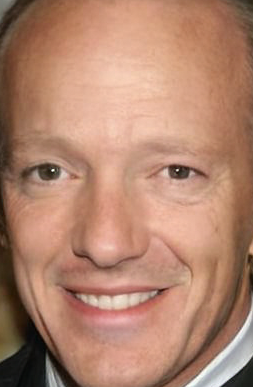}
    \end{minipage}
    \caption{Input preprocessing pipeline: isolating the facial Region of Interest (ROI) from the absolute image domain to remove background confounders.}
    \label{fig:preprocessing}
\end{figure}

Before feature extraction, we isolate the facial ROI from each image as shown in Figure \ref{fig:preprocessing}. This preprocessing eliminates background confounders, ensuring downstream models assess authentic facial synthesis anomalies rather than relying on global environmental shortcuts.

\subsection{Evaluated Foundation Model Paradigms}
We evaluate three distinct structural configurations to capture the trade-offs between training objectives, dataset scale, and model capacity. Table \ref{tab:model_specs} highlights the exact parameter sizes, architecture baselines, and feature dimensions of the models used in this study.

\subsubsection{Supervised Macro-Semantic Paradigm}
To analyze fully supervised classification structures, we evaluate a RoPE-ViT architecture pre-trained on ImageNet-1k (\texttt{vit\_base\_patch16\_rope\_mixed\_ape\_224.naver \_in1k}). The input image $I$ is resized to $224 \times 224 \times 3$, and mapped into a latent vector $f_{s} \in \mathbb{R}^{768}$. Because its pre-training objective requires optimizing hard semantic class boundaries, this representation forces the network to drop ambient variations and prioritize global object symmetry, serving as a baseline for classification-driven features.

\subsubsection{Self-Supervised Geometric Paradigm}
To study unannotated visual alignment, we assess Meta's DINOv3 model pre-trained on the natural web image collection LVD-1689M (\texttt{vit\_large\_patch16\_dinov3.lvd1689m}). Input images are mapped via masked image modeling objectives into a dense latent space $f_{d} \in \mathbb{R}^{1024}$. Lacking explicit categories, DINOv3 preserves localized spatial relationships, rendering it highly sensitive to architectural symmetry and lighting field continuity across the facial surface.

\subsubsection{Agglomerative Multi-Teacher Paradigm}
Representing consolidated foundation models, we evaluate NVIDIA's C-RADIOv4-H. This massive architecture distills multiple teachers simultaneously: geometric tokens from DINOv3, semantic text alignments from SigLIP2, and explicit edge boundaries from SAM3. It maps the input into a high-capacity feature space $f_{r} \in \mathbb{R}^{1280}$, yielding an optimized environment for evaluating multi-task feature spaces.

\subsection{Downstream Linear Probing Configuration}
For any selected foundation model backbone $\mathcal{B}_{\theta}$ parameterized by frozen weights $\theta$, the feature vector is extracted as $f = \mathcal{B}_{\theta}(I)$. We train a linear probe layer parameterized by a weight matrix $W$ and bias $b$ to map the representation to a binary authenticity scalar:
\begin{equation}
    \hat{y} = \sigma\left(W f + b\right)
\end{equation}
where $\sigma(\cdot)$ denotes the standard sigmoid activation function. The optimization optimizes a Binary Cross-Entropy loss function under our data-rich full set configuration.

\begin{table*}[htbp]
\centering
\caption{Architectural specifications and feature dimensions of the evaluated vision foundation model backbones.}
\label{tab:model_specs}
\begin{tabular}{llcc}
\toprule
\textbf{Model Paradigm} & \textbf{Base Architecture} & \textbf{Parameters} & \textbf{Feature Dim ($f$)} \\
\midrule
Supervised ViT & RoPE-ViT-Base/16 & $\sim$86M & 768 \\
DINOv3 & ViT-Large/16 & $\sim$303M & 1024 \\
NVIDIA RADIOv4 & C-RADIOv4-H & $\sim$653M & 1280 \\
\bottomrule
\end{tabular}
\end{table*}

\begin{table*}[ht]
\centering
\caption{Performance comparison on the test set. ViT\_LP, DINOv3\_LP and RADIOv4\_LP represent downstream linear probes optimized on fully trained frozen foundation features across custom trainig set.}
\begin{tabular}{lcccc}
\toprule
\textbf{Model} & \textbf{Precision} & \textbf{Recall} & \textbf{F1 Score} & \textbf{Accuracy} \\
ViT\_LP -Supervised ViT \cite{heo2024rotary}& 0.9118 & 0.9289 & 0.9203 & 0.9086 \\
DINOv3\_LP \cite{simeoni2025dinov3} & 0.9910 & 0.9950 & 0.9930 & 0.9920 \\
RADIOv4\_LP -NVIDIA RADIOv4 \cite{ranzinger2026c} & 0.8965 & 0.8076 & 0.8497 & 0.8378 \\
InceptionResNet \cite{cao2018vggface2} & 0.9869 & 0.9849 & 0.9859 & 0.9840 \\
FreqNet \cite{tan2024frequency} & 0.9936 & 0.9856 & 0.9896 & 0.9882 \\
EfficientNet-B0 & 0.9866 & 0.9886 & 0.9876 & 0.9859 \\
EfficientNet-B2 & 0.9878 & 0.9501 & 0.9686 & 0.9650 \\
BNN \cite{Lanzino_2024_CVPR} & 0.9594 & 0.9749 & 0.9671 & 0.9623 \\
\bottomrule
\end{tabular}
\label{tab:indistribution}
\end{table*}

\section{Experiments}
\label{sec:experiments}

Our experimental protocol evaluates two fundamental performance domains: \textit{in-distribution classification performance} and \textit{out-of-distribution (OOD) generalization}. We organize our benchmarks to measure how structural feature embeddings respond when evaluated on the DF40 benchmark.

\subsection{Dataset Composition}
Our training dataset is built from a diverse multi-source configuration comprising approximately 21,000 authentic and 20,000 manipulated face images. Real samples are collected from CelebA-HQ ($n=10,000$), FFHQ ($n=9,000$), and the LaPa dataset ($n=1,970$). The corresponding forged samples are split into:
\begin{itemize}
    \item \textbf{Entire Face Synthesis:} 10,000 global images from the 100KFake repository and 9,000 images from ThisPersonDoesNotExist, both generated via global GAN architectures.
    \item \textbf{Identity Manipulation:} 1,240 localized structural face-swaps mapped over the LaPa framework.
\end{itemize}
The internal test set maps this layout, containing 1,000 real images from FFHQ and 1,950 from LaPa, paired with manipulated images from ThisPersonDoesNotExist ($n=1,000$) and face-swapped LaPa ($n=1,247$).

\subsection{Evaluation Baselines and Reference Models}
We benchmark the foundation paradigms against several standard reference baselines:
\begin{itemize}
    \item \textbf{CNN Reference Models:} Optimized convolutional models including EfficientNet-B0 and EfficientNet-B2, alongside an InceptionResNetV1 structure initialized with VGGFace2 weights.
    \item \textbf{Forensic-Specific Baselines:} FreqNet \cite{tan2024frequency} and the compressed Binary Neural Network (BNN) forensic model proposed by Lanzino et al. \cite{Lanzino_2024_CVPR}.
\end{itemize}

\subsection{In-Distribution Results}
Table \ref{tab:indistribution} represents the performance comparison for different methods on our test set representing in-distribution results, and it is seen that most of the models work well under in-distribution data condition. 

The clear winner in this context depends on the metric: \textbf{FreqNet} achieves the highest precision overall ($0.9936$), indicating an exceptionally low false-positive rate. However, for balanced operational success, Meta's \textbf{DINOv3} yields the highest comprehensive performance, achieving an F1-Score of $0.9930$ and an overall accuracy of $0.9920$. This confirms that when train and test distributions match, both explicit local frequency fingerprints and massive self-supervised geometric feature spaces can effectively map deepfake authenticity.

\subsection{Cross-Domain Generalization on Benchmark Manifolds}
To analyze true cross-domain generalization, models are evaluated on unseen generative domains from the DF40 benchmark suite. Table \ref{tab:collabdiff_styleclip} isolates structural performance tracking variations on the out-of-distribution CollabDiff and StyleCLIP \cite{patashnik2021styleclip} datasets, which represent specialized face editing deepfake techniques. 

\begin{table*}[htbp]
\centering
\caption{Performance Metrics for DF40: CollabDiff and StyleCLIP Datasets (Both Face Editing Deepfakes).}
\label{tab:collabdiff_styleclip}
\begin{tabular}{lcccccccc}
\toprule
& \multicolumn{4}{c}{\textbf{CollabDiff}} & \multicolumn{4}{c}{\textbf{StyleCLIP}} \\
\cmidrule(r){2-5} \cmidrule(l){6-9}
\textbf{Model} & \textbf{P} & \textbf{R} & \textbf{F1} & \textbf{Acc} & \textbf{P} & \textbf{R} & \textbf{F1} & \textbf{Acc} \\
\midrule
ViT\_LP & 0.5038 & 0.9990 & 0.6698 & 0.5075 & 0.3165 & 0.9980 & 0.4806 & 0.4363 \\
DINOv3\_LP & 0.5165 & 0.9990 & 0.6810 & 0.5320 & 0.2766 & 0.9940 & 0.4328 & 0.3191 \\
RADIOv4\_LP & 0.5445 & 0.9420 & 0.6901 & 0.5770 & 0.4168 & 0.9429 & \textbf{0.5781} & \textbf{0.6403} \\
\midrule
InceptionResNet & 0.3377 & 0.5100 & 0.4064 & 0.2550 & 0.1541 & 0.5145 & 0.2371 & 0.1350 \\
FreqNet & 0.7904 & 0.9920 & \textbf{0.8798} & \textbf{0.8645} & 0.2607 & 0.9970 & 0.4134 & 0.2605 \\
EfficientNet-B0 & 0.5000 & 1.0000 & 0.6667 & 0.5000 & 0.2608 & 0.9970 & 0.4134 & 0.2608 \\
EfficientNet-B2 & 0.4701 & 0.8870 & 0.6145 & 0.4435 & 0.2382 & 0.8819 & 0.3751 & 0.2323 \\
BNN & 0.5015 & 0.9930 & 0.6664 & 0.5030 & 0.2770 & 0.9930 & 0.4332 & 0.3210 \\
\bottomrule
\end{tabular}
\end{table*}

A critical finding emerges when assessing the \textbf{CollabDiff} results: almost all baseline architectures and foundation models experience severe functional degradation. Standard linear probes (ViT\_LP, DINOv3\_LP) and standard CNNs (EfficientNet-B0) converge to an accuracy of approximately $0.5000$, with recall tracking near $1.0000$. This specific mathematical distribution indicates total model collapse; the classifiers are failing to map meaningful representations and are instead reverting to a trivial random-guessing shortcut by predicting all inputs as fake. Similarly for StyleCLIP results.

A primary driver behind this performance failure is the intrinsic data quality of the CollabDiff and StyleCLIP split, where source real images exhibit a very low native patch resolution ($\approx 90 \times 120$ pixels). When these low-resolution tensors are upscaled to satisfy model requirements, micro-textural forensic boundaries are heavily degraded or entirely obliterated. In this degraded setting, standard models fail. However, \textbf{FreqNet} bypasses this limitation, acting as the winner for CollabDiff ($0.8645$ Accuracy) due to its specialized frequency tracking network. Crucially, when evaluating the more complex \textbf{StyleCLIP} pipeline, FreqNet collapses ($0.2605$ Accuracy), positioning \textbf{NVIDIA RADIOv4} as the clear multi-domain generalization winner, maintaining an Accuracy of $0.6403$ and F1-Score of $0.5781$ by leveraging its multi-teacher edge and segmentation tokens.

Table \ref{tab:midjourney_whichface} represents performance comparison for MidJourney \cite{midjourney2024} and WhichFaceIsReal \cite{whichfaceisreal}, which emphasize entire face synthesis deepfake challenges. This isolated grouping facilitates an unconfounded assessment of model behavior across disparate generative architectures.

\begin{table*}[htbp]
\centering
\caption{Performance metrics for DF40: MidJourney and WhichFaceIsReal datasets (Both Entire Face Synthesis Deepfakes).}
\label{tab:midjourney_whichface}
\begin{tabular}{lcccccccc}
\toprule
& \multicolumn{4}{c}{\textbf{MidJourney}} & \multicolumn{4}{c}{\textbf{Whichfaceisreal}} \\
\cmidrule(r){2-5} \cmidrule(l){6-9}
\textbf{Model} & \textbf{P} & \textbf{R} & \textbf{F1} & \textbf{Acc} & \textbf{P} & \textbf{R} & \textbf{F1} & \textbf{Acc} \\
\midrule
ViT\_LP & 0.9852 & 1.0000 & \textbf{0.9926} & \textbf{0.9907} & 0.9398 & 0.5780 & 0.7158 & 0.7706 \\
DINOv3\_LP & 0.8490 & 0.9950 & 0.9162 & 0.8877 & 0.9833 & 0.8250 & \textbf{0.8972} & \textbf{0.9055} \\
RADIOv4\_LP & 0.9265 & 0.9710 & 0.9482 & 0.9346 & 0.9234 & 0.4820 & 0.6334 & 0.7211 \\
\midrule
InceptionResNet & 1.0000 & 0.9850 & 0.9924 & \textbf{0.9907} & 0.9623 & 0.3570 & 0.5208 & 0.6717 \\
FreqNet & 0.8324 & 0.9980 & 0.9077 & 0.8748 & 0.7759 & 0.9350 & 0.8481 & 0.8326 \\
EfficientNet-B0 & 0.8460 & 0.9940 & 0.9140 & 0.8846 & 1.0000 & 0.4410 & 0.6121 & 0.7206 \\
EfficientNet-B2 & 0.9327 & 0.9980 & 0.9643 & 0.9543 & 1.0000 & 0.5590 & 0.7171 & 0.7796 \\
BNN & 0.9205 & 0.9500 & 0.9350 & 0.9186 & 0.9643 & 0.7020 & 0.8125 & 0.8381 \\
\bottomrule
\end{tabular}
\end{table*}

For entire image generation challenges, the feature distribution changes. Because full synthesis leaves global geometric markers, standard visual feature layers achieve strong performance. For MidJourney, the \textbf{Supervised ViT} configuration functions flawlessly, tying with InceptionResNet at a peak accuracy of $0.9907$. For WhichFaceIsReal, Meta's \textbf{DINOv3} acts as the clear, decisive winner, reaching an accuracy of $0.9055$ and a leading F1-Score of $0.8972$, outperforming both supervised setups and the heavier multi-teacher layouts.

\section{Conclusion}
\label{sec:conclusion}

In this work, we presented a rigorous empirical study mapping the cross-domain generalization limits of Vision Foundation Models (VFMs) for facial deepfake detection on the DF40 benchmark. By freezing the parameters of modern visual backbones—representing supervised, self-supervised, and multi-teacher learning paradigms—and training downstream linear probes, we isolated the pure descriptive capacity of their latent feature spaces. 

Our findings demonstrate a stark divergence in performance based on the underlying forgery mechanism. While frozen foundational features easily capture global structural deformations in entire face synthesis challenges (e.g., MidJourney), they experience significant performance degradation when confronted with localized face editing techniques (e.g., StyleCLIP). Among the evaluated paradigms, the agglomerative multi-teacher representation (\mbox{NVIDIA C-RADIOv4-H}) emerged as the most resilient baseline under extreme domain shifts, successfully retaining edge and semantic boundaries where traditional CNNs and standard self-supervised models collapsed to near-random guessing. This underscores the critical value of multi-task pre-training objectives in generating robust, general-purpose forensic descriptors.

\section{Limitations and Future Work}
\label{sec:limitations}

Despite the strong foundational baselines established in this study, several critical limitations remain that pave the way for future research directions:

\begin{itemize}
    \item \textbf{Sensitivity to Underlying Grid Resolution:} As observed in our out-of-distribution evaluation on CollabDiff, downstream linear probes suffer severe representation failure when evaluating inputs with low native spatial dimensions (e.g., $90 \times 120$). This structural limitation indicates that interpolation and upsampling artifacts can confound standard visual token layers, limiting out-of-the-box model application on low-quality media pipelines.
    \item \textbf{Loss of Localized Spatial Inconsistencies:} Our downstream framework relies on a linear probe optimized over globally pooled token representations. While global average pooling yields compact vectors suitable for linear probing, it fundamentally discards fine-grained spatial relationships and localized patch-level inconsistencies. This structural bottleneck explains why models failed to robustly track micro-blending artifacts in localized face editing datasets.
    \item \textbf{Absence of Cross-Patch Frequency Modeling:} While frequency-aware networks like FreqNet excel at recognizing specific upsampling periodicities, our foundation backbones operate entirely within the spatial domain. The current linear probing configuration lacks an explicit mechanism to bridge macro-structural geometric alignment with pixel-level spectral artifacts.
\end{itemize}

To address these limitations, future work will pivot away from global pooling strategies. We intend to explore token-level consistency networks that explicitly measure the mutual information or structural coherence between individual spatial patches. Furthermore, integrating cross-attention mechanisms that combine spatial foundation features with learnable local frequency descriptors represents a promising frontier for achieving truly universal, domain-agnostic deepfake detection.

\bibliographystyle{unsrt}
\bibliography{sn-bibliography}

\end{document}